\documentclass{article} 
\usepackage{iclr2023_conference,times}

\usepackage{amsmath,amsfonts,bm}

\def\eqref#1{equation~\ref{#1}}

\def\1{\bm{1}}

\DeclareMathAlphabet{\mathsfit}{\encodingdefault}{\sfdefault}{m}{sl}
\SetMathAlphabet{\mathsfit}{bold}{\encodingdefault}{\sfdefault}{bx}{n}

\usepackage{hyperref}
\usepackage{url}
\usepackage{graphicx} 
\graphicspath{ {./images/} }
\usepackage{wrapfig}

\title{SLGTformer: An Attention-Based Approach \\ to Sign Language Recognition}

\author{Neil Song \\
St. Mark's School of Texas\\
Dallas, TX 75230, USA \\
\And
Yu Xiang (Advisor) \\
Department of Computer Science \\
University of Texas at Dallas \\
Dallas, TX 75080, USA \\
}

\iclrfinalcopy 
\begin{document}

\maketitle

\begin{abstract}
Sign language is the preferred method of communication of deaf or mute people, but similar to any 
language, it is difficult to learn and represents a significant barrier for those who are hard of 
hearing or unable to speak. A person’s entire frontal appearance dictates and conveys specific 
meaning. However, this frontal appearance can be quantified as a temporal sequence of human body 
pose, leading to Sign Language Recognition through the learning of spatiotemporal dynamics of 
skeleton keypoints. We propose a novel, attention-based approach to Sign Language Recognition 
exclusively built upon decoupled graph and temporal self-attention: the Sign Language Graph Time 
Transformer (SLGTformer). SLGTformer first deconstructs spatiotemporal pose sequences separately into spatial graphs and temporal windows. SLGTformer then leverages novel Learnable Graph Relative Positional Encodings (LGRPE) to guide spatial self-attention with the graph neighborhood context of the human skeleton. By modeling the temporal dimension as intra- and inter-window dynamics, we introduce Temporal Twin Self-Attention (TTSA) as the combination of locally-grouped temporal attention (LTA) and global sub-sampled temporal attention (GSTA). We demonstrate the effectiveness of SLGTformer on 
the World-Level American Sign Language (WLASL) dataset, achieving state-of-the-art performance 
with an ensemble-free approach on the keypoint modality. The code is available at \url{https://github.com/neilsong/slt}
\end{abstract}

\section{Introduction}

Sign language is, at its core, a visual language combining the varied and nuanced cues of body 
orientation, hand gesture, and facial expression, primarily used as a means of communication by 
the deaf and/or mute community. However, sign language is difficult to learn, presenting the same 
problem that foreign spoken languages do to native speakers. Furthermore, sign language is 
language-specific and has evolved dialects over time. With over 300 sign languages worldwide, even
intralingual communication and understanding is severely limited \citep{lucas_2001}. To assist in 
the facilitation of communication between spoken language users and sign language users, Sign 
Language Recognition (SLR) has emerged as an area of application for advanced computer vision 
research. Since then, there have been a variety of systems and solutions proposed to tackle this 
problem.

Traditional methods utilize conventional classifiers e.g. SVM \citep{8666491, 8126796, doi:10.1080/01691864.2018.1490666}, k-means \citep{Wei2016-ki}, KNN \citep{doi:10.1080/01691864.2018.1490666}, and decision trees \citep{7464244}. Although traditional 
methods provided high accuracy rates, they often used wearable or glove sensor data as input 
modalities. Furthermore, they lacked the generalization and temporal expressiveness of modern deep
learning methods, so vision-based deep learning approaches began to rise in popularity. Initially,
3DCNN, LSTM, and 3DCNN-LSTM were introduced as methods of SLR \citep{9078786, 9423424, 9210578, 9158332}. Early attempts at incorporating attention mechanisms to model 
temporal dynamics were also introduced \citep{10.5555/3504035.3504310, 8466903, sam_slr_v2}. Although not yet applied in sign language recognition, space-time divided 
global transformer attention in TimeSformer \citep{timesformer} as an extension of the Vision Transformer \citep{vit} 
has been shown to be empirically superior to both traditional convolution and 
RNN or LSTM. However, in general, these methods are not sufficient or robust enough to learn the 
nuanced pose and spatial dynamics of sign language for datasets that are limited in annotations 
and expansive in vocabulary size. Furthermore, the entire RGB frame contains very little semantic 
information outside of segmentation of the signer, ultimately resulting in inefficient utilization
of computation and memory resources.

A trend in these early methods is that they all do not yet exploit the inherent spatial locality 
information implicitly available in the construction of the human skeleton. As all subjects are 
humans, skeleton keypoint location and human body pose can be used as structural guidance along 
potentially both spatial and temporal dimensions, a vast improvement to the general spatial and 
temporal modeling techniques used in earlier approaches. Building off recent advancements in 
action recognition, methods using GNN, GCN, and ST-GCN have attracted much research attention 
\citep{WANG2021211, 8954298, 10.1145/1557019.1557109, https://doi.org/10.48550/arxiv.1904.12659, 8954160, Shi_2020, Song_2020}. Recently, further improvements have been made in the graph deep learning domain 
that have yet to appear in action recognition, such as the introduction of graph transformer attention, 
positional encodings, and structural encodings \citep{https://doi.org/10.48550/arxiv.2012.09699, NEURIPS2021_b4fd1d2c, ying2021do, https://doi.org/10.48550/arxiv.2106.05667, NEURIPS2021_6e67691b, graphgps}. Although only proven to be empirically effective and
superior on either molecular or image datasets, the generalization capacity of both graph 
structures and global transformer attention includes minor domain specialization similar to the 
action recognition application of the GCN. In general, the skeleton keypoint modality expresses 
the majority of semantic meaning presented by the signer's frontal appearance, as nonmanual cues 
are of relatively less importance.

Although existing skeleton-based methods offer great improvements to the performance of action 
recognition and the basic subtask of sign language recognition, it requires additional annotation in 
the form of keypoints. However, this presents not only a higher logistical burden for sign 
language dataset creation but also a problem for existing sign language datasets without keypoint 
annotation. Even those with keypoint annotation often lack the joints of the hand, a result of 
using ground-truth Kinect data \citep{9210578, li2020word, forster-etal-2012-rwth, Albanie2020bsl1k}. Our solution for the lack of keypoint annotation follows 
\citet{sam_slr_v2}: using a pre-trained wholebody pose estimator \citep{Zhang_2020_CVPR, Sun_2019_CVPR} as a vision-based keypoint estimation approach. Although this approach 
provides inherently inaccurate data, the estimator outputs a score value for every keypoint, 
allowing the learning of relative prominence of each keypoint to counterbalance this effect.

We introduce Sign Language Graph Time Transformer (SLGTformer) as a solution to the limited 
spatial expression of the RGB-based methods and the underperformant temporal expression of RNN and
LSTM. By using global transformer attention after added neighborhood context and latent graph 
structure by learnable graph relative positional encodings (LGRPE), 
SLGTformer fully leverages the semantic expression of the skeleton keypoint and graph modality. LGRPE is first initialized from the given skeleton graph structure before SLGTformer learns latent representations of ambiguous relative positions of nodes.
Furthermore, to enhance the modeling of temporal dynamics, SLGTformer also implements locally-grouped temporal attention (LTA) and
global sub-sampled temporal attention (GSTA), jointly dubbed temporal twin self-attention (TTSA). This allows temporal
dependencies to be modeled both within and across local windows 
efficiently, inspired by the spatially separable self-attention
(SSSA) of Twin-SVT \citep{chu2021Twins}. By decoupling the two attention mechanisms into separate 
dimensions, we allow each weight matrix to learn and attend separately to two non-orthogonal 
latent spaces. Relative to current skeleton-based methods in SLR, SLGTformer presents a radically different attention-based architecture and design and represents the first attempt at a graph-informed, purely attention based approach to SLR.

In summary, the key contributions of this study are as follows:
\begin{itemize}
 \item A novel, graph-time decoupled attention schema transformer, namely, SLGTformer, is 
 introduced for SLR.
 \item Novel graph-initialized learnable graph relative positional encodings (LGRPE) are introduced to better encode graph structures in spatial attention.
 \item Novel one-dimensional temporal twin self-attention (TTSA) is introduced 
 to efficiently model short-range and long-range temporal dependencies.  
\end{itemize}

\section{Related Work}
In this section, relevant studies are reviewed and discussed from several perspectives: existing SLR 
methods, graph neural networks, and vision transformers. SLGTformer builds upon the many insights discovered within these respective fields as relevant to an attention-based approach to keypoint sequence recognition.

\subsection{Sign Language Recognition}
Significant progress has been made in Sign Language Recognition (SLR) with the acceleration of recent developments in deep learning computation and architecture \citep{9078786, 9423424, 9210578, 9158332}. The spatiotemporal dynamics of 
sign language were modeled using two main approaches with the convolution mechanism at the core. 
The CNN and RNN (LSTM) were combined as an approach to SLR in \citep{8458185, 8806467}. An 
alternative approach to modeling temporal dynamics was the 3DCNN, using 3D convolutions to embed 
temporal semantics in latent features \citep{8466903, 9423424}. These approaches are all 
inspired by and improvements upon similar ideas and architectures seen in action recognition 
\citep{8026287, 8099985}. Several multi-cue fusion networks have been proposed to leverage both 
manual and nonmanual cues with methods such as hand pose priors and late fusion multi-modal 
ensemble \citep{9423424, Hu_Zhou_Li_2021, sam_slr_v2}. Multi-cue methods use
upper-body pose, facial features, and hand pose to improve the recognition performance in SLR. Although these methods are certainly robust and effective, they require several additional modalities and significantly higher amounts of input data, limiting not only the scope of their real-life deployment but also the overall effectiveness of the solution outside the sandbox of research experiments. 

\subsection{Graph Neural Networks}

Works in graph neural networks first started gaining mainstream traction in the SLR-adjacent action recognition field with the introduction of Spatial Temporal Graph Convolutional Networks \citep{Yan_Xiong_Lin_2018}. Since this major advancement, various graph convolutional network improvements were made such as decuopled graph convolutions and graph-based dropout layers \citep{cheng2020eccv} as well as multi-granular \citep{10.1145/3474085.3475574} and topology approaches \citep{ye2020dynamic}. However, attention-based approaches have also been making advancements in more general graph-based areas such as molecular analysis. GraphGPS uses general recipe combinations of several graph positional and structural encodings such as Laplacian Positional Encodings and Random Walk Structural Encodings along with different Message Passing Neural Networks to better guide global attention \citep{graphgps}. Earlier approaches such as Graphormer only used structural encodings or different variations of positional encodings \citep{ying2021do}. Notably, \citet{park2022grpe} recently introduced graph relative positional encodings. However, they solve the ambiguity of relative node position by representing topology encodings as distinct dot products with query and key matrices. In SLGTformer, we take the approach of learnable, higher-dimension relative positional encodings, using a small meta-network to expand topological generalization capacity. 

\subsection{Vision Transformers}

Recently, there has been an explosion of convolution-free approaches in computer vision, with global attention across image patches demonstrating impressive generalization capacity and competitive performance \citep{vit}. Windowed local-global attention schemas such as Swin's shifted windows approach and Twin-SVT's spatially separable self-attention provide significant improvements in ViT performance \citep{liu2021Swin, liu2021swinv2, chu2021Twins}. ConViT's convolution kernel-like relative positional encoding initialization also demonstrates the effectiveness of strong initializations. \citep{d2021convit}. With the rise of ViT, attention-based models have also seen active development in the action recognition field. Divided space-time attention in TimeSformer \citep{timesformer} separates global attention along spatial and temporal dimensions and demonstrates proficiency on several datasets \citep{8099985, smtgsmtgv2, miech19howto100m}.

\begin{figure}[ht]
    \centering
    \includegraphics[width=0.92\textwidth]{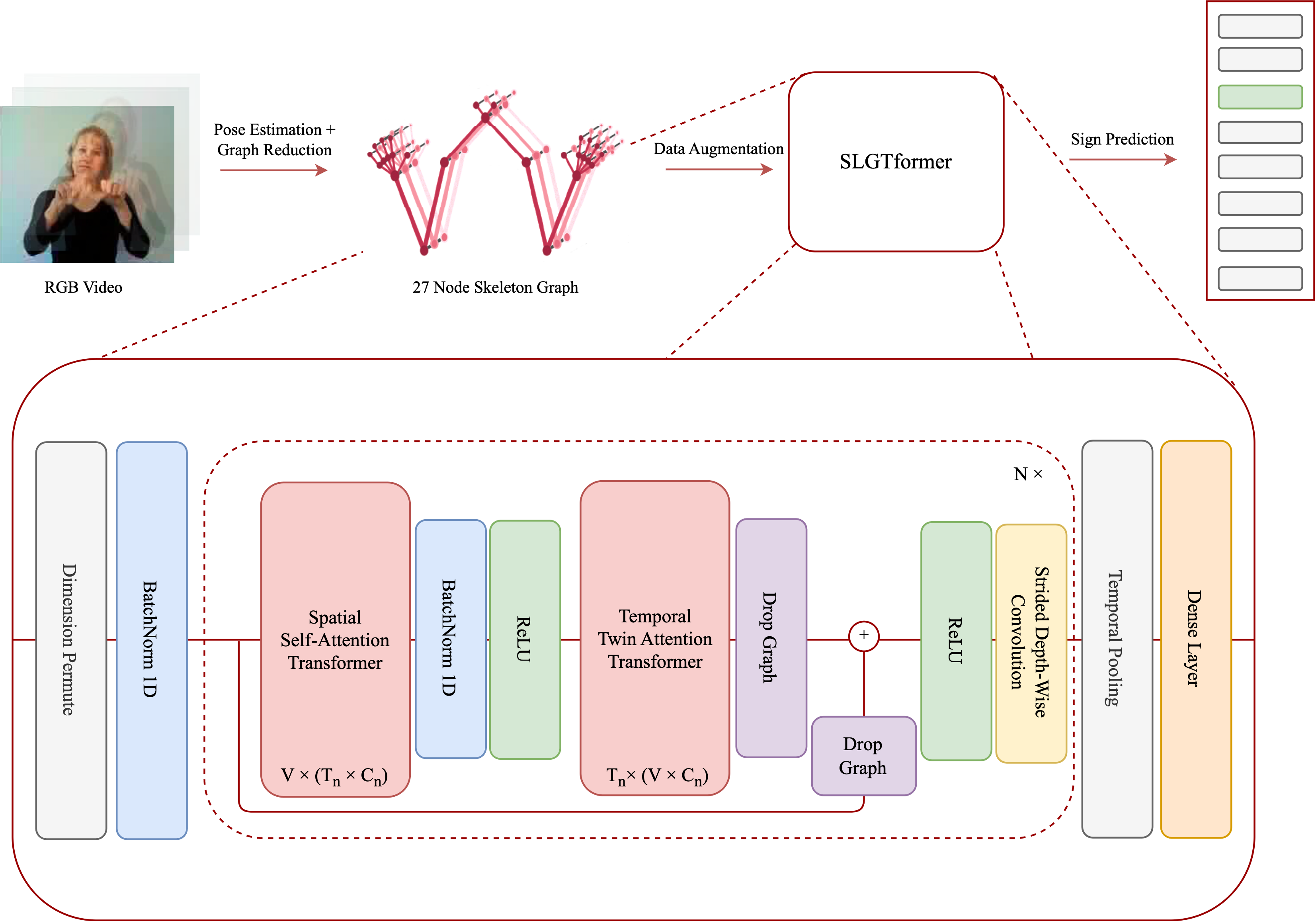}
    \caption{Illustration of the SLGTformer pipline. Starting from an RGB video, pose keypoints are estimated and graph reduction is performed to obtain the input keypoint sequence. The sign label is predicted from the network with $N$ repeating decoupled spatial and temporal attention transformers.}
    \label{fig:overview}
\end{figure}

\section{Methodology}
This section describes the details of the SLGTformer pipeline as illustrated in \ref{fig:overview}. Given an input RGB video $X \in$ $\mathbb{R}^{H \times W \times 3 \times T}$ consisting of $T$ RGB frames of size $H \times W$, we estimate keypoints offline to obtain input keypoint sequence $S_0 \in$ $\mathbb{R}^{N \times 2 \times T}$, where $N$ denotes number of joints per frame. All dimension permutations and channel projections are applied before attention layers by $P\colon S_0 \to S \in$ $\mathbb{R}^{T \times N \times D_{emb}}$.

\subsection{Spatial Self-Attention}
In SLGTformer spatial self-attention, we use novel Learnable Graph Relative Positional Encodings (LGRPE) and the spatial-temporal convolution factor under the spatial partitioning strategy with decoupled adjacency matrices \citep{sam_slr_v2, Yan_Xiong_Lin_2018}. 

\subsubsection{Learnable Graph Relative Positional Encodings}

We define the initialization of the LGRPE as
\begin{equation}
    \Gamma_{(i,j)}^0 = \psi\left(i,j\right),
\end{equation}
where $\psi(i,j)$ is the shortest path distance between nodes $i$ and $j$ and $\Gamma \in$ $\mathbb{R}^{N^2}$. The effectiveness of using initialization to encode known positional context has been demonstrated in \citet{d2021convit, park2022grpe}. 

\begin{wrapfigure}{r}{0.53\textwidth}
\includegraphics[width=0.53\textwidth]{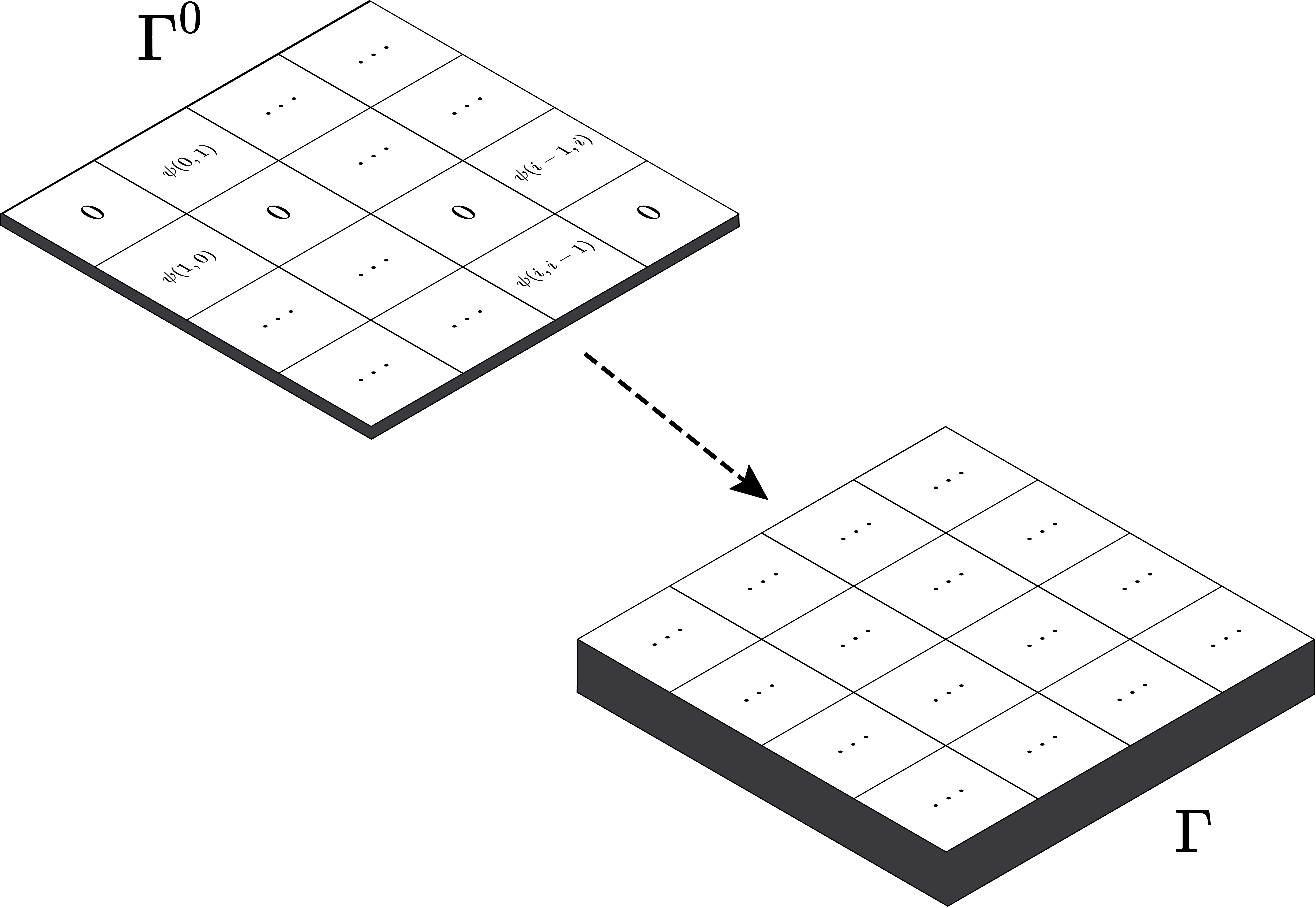}
\centering
\caption{Meta-network learns unique positional encodings $\Gamma$ from $\Gamma^0$ where $\psi$ denotes shortest path distance.}
\end{wrapfigure}

However, with $\psi$ defining the relative position of node $i$ to node $j$, there will exist relative positional ambiguity between node $h$ and node $j$ when $\psi(i, h) = \psi(i, j)$. 
Thus, we allow SLGTformer to learn representation $\Gamma \in$ $\mathbb{R}^{N^2 \times C_{pos}}$ from $\Gamma^0$ with a small meta-network, effectively using both higher dimensionality and gradient descent to resolve the ambiguity of $\psi$. The meta-network is a 2-layer MLP with a
ReLU activation in between, defined as
\begin{equation}
    \Gamma_{(i,j)} = \operatorname{MLP}\left(\Gamma^0_{(i,j)}\right).
\end{equation}

\subsubsection{Multi-Head Self-Attention}
We begin with an introduction to the basics of self-attention (SA) layers before introducing positional self-attention (PSA), the chosen SA variant used to inject the graph neighborhood context through LGRPE \citep{10.5555/3454287.3454294}.

We define spatial dot product attention as the trainable associative memory between the vector pair (key, query). A series of $N$ ``query'' embeddings denoted as $Q \in \mathbb{R}^{N \times D_h}$ is matched with another series of $N$ ``key'' embeddings denoted as $K \in \mathbb{R}^{N \times D_h}$. In self-attention, the sequence is matched to itself, so we define $Q$, $K$, $V$ as the following:
\begin{equation}
    \mathbf{Q}^{(a)}_{(n)} = \mathbf{W}^{(a)}_Q \operatorname{LN}\left(\mathbf{S}_{S}^{(n)}\right) \in \mathbb{R}^{D_h},
\end{equation}
\begin{equation}
    \mathbf{K}^{(a)}_{(n)} = \mathbf{W}^{(a)}_K \operatorname{LN}\left(\mathbf{S}_{S}^{(n)}\right) \in \mathbb{R}^{D_h},
\end{equation}
\begin{equation}
    \mathbf{V}^{(a)}_{(n)} = \mathbf{W}^{(a)}_V \operatorname{LN}\left(\mathbf{S}_{S}^{(n)}\right) \in \mathbb{R}^{D_h},
\end{equation}
where LN denotes LayerNorm \citep{DBLP:journals/corr/BaKH16}, $a = 1,\ldots,\mathcal{A}$ is an index over all attention heads, $\mathbf{S}_S \in \mathbb{R}^{N \times D_{emb}}$ is batched along the temporal dimension $(B \times T)$, $\mathbf{W}_Q \in \mathbb{R}^{D_{emb} \times D_{h}}$, $\mathbf{W}_K \in \mathbb{R}^{D_{emb} \times D_{h}}$, and $\mathbf{W}_V \in \mathbb{R}^{D_{emb} \times D_{h}}$. The resulting vanilla self-attention map, depicts the ``semantic similarity" between $\mathbf{Q}_i$ and $\mathbf{K}_j$:
\begin{equation}
\label{eqn:dotproduct}
    \boldsymbol{\alpha}_{(n)}^{(a)}=\operatorname{SM}\left(\frac{{\mathbf{Q}_{(n)}^{(a)}}^\top}{\sqrt{D_{h}}} \cdot\left[\mathbf{K}_{(0)}^{(a)}\left\{\mathbf{K}_{\left(n^{\prime}\right)}^{(a)}\right\}_{\substack{n^{\prime}=1, \ldots, N}}\right]\right),
\end{equation}
where SM denotes the softmax activation function. Then, we compute the weighted sum of the value vectors, $\mathbf{V}$, with $\boldsymbol{\alpha}^{(a)}$ coeffecients.
\begin{equation}
    \mathbf{s}_{(n)}^{(a)}=\alpha_{(n,0)}^{(a)} \mathbf{V}_{(0)}^{(a)}+\sum_{n^{\prime}=1}^{N} \alpha_{(n,n^{\prime})}^{(a)} \mathbf{V}_{\left(n^{\prime}\right)}^{(a)}.
\end{equation}

Lastly, the concatenation of these vectors from all heads
is projected and passed through an MLP, using residual
connections from previous block, $b-1$, after each operation:

\begin{equation}
    \mathbf{S}_S^{\prime(n, b)} = W_{O}\left[\begin{array}{c}
    \mathbf{s}_{(n)}^{(1)} \\
    \vdots \\
    \mathbf{s}_{(n)}^{( \mathcal{A})}
    \end{array}\right]+\mathbf{S}_S^{(n, b-1)},
\end{equation}
\begin{equation}
    \mathbf{S}_S^{(n,b)} = \operatorname{MLP}\left(\operatorname{LN}\left(\mathbf{S}_S^{\prime(n,b)}\right)\right)+\mathbf{S}_S^{\prime(n,b)}.
\end{equation}

\subsubsection{Positional Self-Attention}
However, since we use LGRPE, we modify \ref{eqn:dotproduct} to add neighborhood positional context following \citet{d2021convit, 10.5555/3454287.3454294}:
\begin{equation}
        \boldsymbol{\alpha}_{(n)}^{(a)}=\operatorname{SM}\left(\frac{{\mathbf{Q}_{(n)}^{(a)}}^\top}{\sqrt{D_{h}}} \cdot\left[\mathbf{K}_{(0)}^{(a)}\left\{\mathbf{K}_{\left(n^{\prime}\right)}^{(a)}\right\}_{\substack{n^{\prime}=1, \ldots, N}}\right] + {v_{pos}^{(a)}}^\top\Gamma_{(n,n^\prime)}^{(a)}\right),
\end{equation}
where each attention head contains trainable embedding $v_{pos}^{(a)} \in \mathbb{R}^{D_{pos}}$.

\subsubsection{Post-Attention Graph Convolution Factor}
A traditional graph convolution is represented as $2D$ convolution multiplied by a factor derived from the adjacency matrix of intra-body connections $\mathbf{A}$ \citep{Yan_Xiong_Lin_2018, sam_slr_v2}:
\begin{equation}
    \mathbf{x_{out}} = \mathbf{D^{-\frac{1}{2}}}\left(\mathbf{A} + \mathbf{I}\right)\mathbf{D^{-\frac{1}{2}}}\mathbf{x_{in}}\mathbf{W_c},
\end{equation}
where $\mathbf{I}$ represents an identity matrix of self-connections, $\mathbf{D}$ stands for the diagonal degree of $\left(\mathbf{A} + \mathbf{I}\right)$, and $\mathbf{W_c}$ is the trainable weights of the convolution.

Multiplying the factor, $\mathbf{D^{-\frac{1}{2}}}\left(\mathbf{A} + \mathbf{I}\right)\mathbf{D^{-\frac{1}{2}}}$, post-attention boosts graph position context directly from the adjacency matrix:

\begin{equation}
    \mathbf{S}_{S,f}^{(n,b)} = \mathbf{D^{-\frac{1}{2}}}\left(\mathbf{A} + \mathbf{I}\right)\mathbf{D^{-\frac{1}{2}}}\mathbf{S}_{S}^{(n,b)}.
\end{equation}

However, following \citet{cheng2020eccv, sam_slr_v2}, we use the decoupled graph convolution factor instead, such that $\widetilde{\mathbf{A}} \in \mathbb{R}^{n \times n \times g}$ is a learnable representation of $\mathbf{D^{-\frac{1}{2}}}\left(\mathbf{A} + \mathbf{I}\right)\mathbf{D^{-\frac{1}{2}}}$ divided into $g$ groups. The final result is the channel-wise concatenation of the multiplied vectors:

\begin{equation}
    \mathbf{S}_{S,f}^{(n,b)} = \left[\begin{array}{c}
    \widetilde{\mathbf{A}}^{(1)}\mathbf{S}_{S}^{(1,b)} \\
    \vdots \\
    \widetilde{\mathbf{A}}^{(g)}\mathbf{S}_{S}^{(g,b)}
    \end{array}\right].
\end{equation}

\begin{figure}[h]
    \centering
    \includegraphics[width=0.63\textwidth]{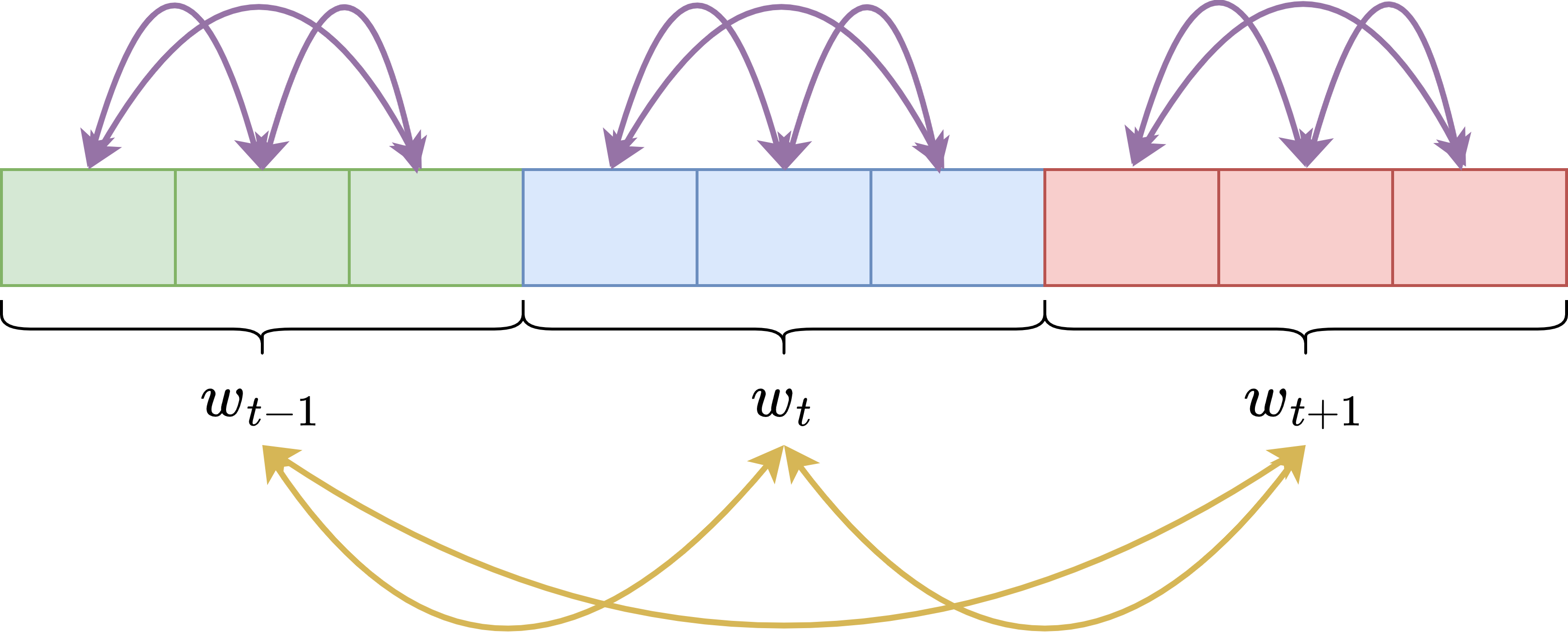}
    \caption{Depiction of intra-window communication in Locally Grouped Temporal Attention (LTA) as purple and inter-window communication in Global Sub-Sampled Temporal Attention (GSTA) as yellow. LTA and GSTA are interleaved in Temporal Twin Self Attention (TTSA).}
    \label{fig:ttsa}
\end{figure}

\subsection{Temporal Twin Self Attention}
Temporal Twin Attention (TTSA) builds upon the paradigms of local-global attention first seen in \citet{chu2021Twins}. Because the temporal dimension no longer faces the same neighborhood positional graph restrictions as the spatial dimension, we treat it analogous to a one-dimensional stream of tokens.

\subsubsection{Locally Grouped Temporal Attention}
We define locally grouped temporal attention (LTA) as self-attention within $\frac{T}{w}$ one-dimensional windows of size $w$. This is similar to the group design of depthwise convolutions and also remains consistent with the multi-headed design of self-attention, where semantic matching only occurs within channels of the same head. The computation cost of one window is $\mathcal{O}\left(\frac{T^2}{w^2}D_h\right)$ and the total cost is $\mathcal{O}\left(\frac{T^2}{w}D_h\right)$. If we consider $k = \frac{T}{w}$, the total computation cost is $\mathcal{O}\left(kTD_h\right)$, which grows linearly with $T$ if $k$ remains constant. Windowed self-attention is analogous to modeling short-range temporal dynamics in the sequence.

However, in the same way that we cannot replace all convolutions in a CNN with depthwise convolutions, there must be communication between local groups. This is implemented through window-shifting in Swin \citep{liu2021Swin, liu2021swinv2}, and following \citet{chu2021Twins}, we use global attention to solve this problem.

\subsubsection{Global Sub-Sampled Temporal Attention}
Although we can use vanilla global SA, as defined in \ref{eqn:dotproduct}, along all temporal tokens, this essentially renders LTA redundant and also increases our time complexity to $\mathcal{O}\left(T^2D_h\right)$. However, we still need a mechanism to model long-range temporal dependencies.

We introduce Global Sub-Sampled Temporal Attention (GSTA) to attend globally to semantic representations of the local groups. Essentially, the sub-sampled feature map of each window is the key in the self-attention mechanism, and if we use LTA and GSTA like depth-wise and point-wise separable convolutions respectively, the total time complexity comes out to $\mathcal{O}\left(\frac{T^2D_h}{k} + kTD_h\right)$. We set $k = 6$ everywhere for simplicity, while the sub-sample window-size is $4, 2, 1$ for the last three stages respectively since later stages have lower resolutions. Following \citet{chu2021Twins}, we use regular strided convolutions as the sub-sampling function.

We can formally write TTSA as the following:

\begin{equation}
\begin{aligned}
&\hat{\mathbf{S}}_{T}^{(t, b)}=\operatorname{LTA}\left(\operatorname{LN}\left(\mathbf{S}_{T}^{(t, b-1)}\right)\right)+\mathbf{S}_{T}^{(t, b-1)}, \\
&\mathbf{S}_{T}^{(t,b)}=\operatorname{MLP}\left(\operatorname{LN}\left(\hat{\mathbf{S}}_{T}^{(t, b)}\right)\right)+\hat{\mathbf{S}}_{T}^{(t, b)}, \\
&\hat{\mathbf{S}}_T^{b+1}=\operatorname{GSTA}\left(\operatorname{LN}\left(\mathbf{S}_T^b\right)\right)+\mathbf{S}_T^b, \\
&\mathbf{S}_T^{b+1}=\mathrm{MLP}\left(\operatorname{LN}\left(\hat{\mathbf{S}}_T^{b+1}\right)\right)+\hat{\mathbf{S}}_T^{b+1}, \\
&t \in\{1,2, \ldots, w\}
\end{aligned}
\end{equation}

where  $\mathbf{S}_T \in \mathbb{R}^{T \times D_{emb}}$ is batched along the spatial dimension $(B \times N)$. Both LTA and GSTA have multiple heads as in standard self-attention. We use PEG of \citet{chu2021conditional} to encode position information after the first block in each stage.

\begin{table}[t]
\begin{center}
\begin{tabular}{ll}
\multicolumn{1}{c}{Variations}  &\multicolumn{1}{c}{Top-1}
\\ \hline \\
\bf SLGTformer         &47.42 \\
w/o LGRPE             &46.76 \\
w/o TTSA            &46.13 \\
w/o Post-Attention Factor &45.31 \\
\end{tabular}
\end{center}
\caption{Ablation Studies on SLGTformer on WLASL2000 test set}
\label{ablation}
\end{table}

\begin{table}[t]
\begin{center}
\begin{tabular}{lll}
\multicolumn{1}{c}{Model}  &\multicolumn{1}{c}{\bf Top-1} &\multicolumn{1}{c}{\bf Top-5}
\\ \hline \\
Baseline*\textsubscript{(R)}             &32.48  &57.31\\
TimeSformer Base*\textsubscript{(R)}    &33.36 &71.02 \\
TimeSformer HR*\textsubscript{(R)}      &43.66 &80.14 \\
SAM-SLR-v2 RGB\textsubscript{(R)}     &47.51 &80.31
\\ \hline
ST-GCN \textsubscript{(K)}          &37.91 &71.26 \\
SAM-SLR-v2 Joint*\textsubscript{(K)}    &44.16 &76.41 \\
SAM-SLR-v2 Joint\textsubscript{(K)}      &45.61 &77.79 \\
\hline
Fusion-3 \textsubscript{(R, K)}           &38.84 &67.58 \\
Hand + RGB \textsubscript{(R,K)}        &51.39 &86.34 \\
\hline
\bf SLGTformer\textsubscript{(K)}          &47.42 & 79.58\\
\end{tabular}
\end{center}
\caption{Performance of SLGTformer on WLASL2000 test set. (R) Used RGB modality. (K) Used keypoint modality. * Results we have reproduced.}
\label{performance}
\end{table}

\section{Experiments}
In this section, we report the experimental results of our proposed SLGTformer on the major sign language recognition
dataset, World-Level American Sign Language (WLASL) \citep{li2020word}. We then demonstrate the effectiveness of
proposed novel approaches via ablation studies on the components of SLGTformer and comparisons with the state-of-the-art. Last, we discuss the limitations of SLGTformer.

\subsection{Dataset}
WLASL2000 Dataset is a American Sign Language with
a vocabulary size of 2000 words. It is a challenging dataset collected from web videos performed by 119 signers. It contains 21,083 samples with unconstrained recording conditions. The dataset is imbalanced, and the average samples per video are low. 

\subsection{Data Preparation}
In spirit of maintaining consistent keypoint data, we follow \citet{sam_slr_v2} in data preparation procedure. Using the MMPose \citep{mmpose2020} a pretrained whole-body pose estimator with an HRNet \citep{HRnet} backbone,  a 27-node 2D graph is constructed. Keypoint coordinates are normalized to [-1,1]. Random
sampling, mirroring, rotating, scaling, jittering, and shifting
are applied as data augmentations. We use a clip duration
of 150 frames, repeating the last frame of videos with lesser than 150 frames. In training, 120 frames are randomly and contiguously sampled from the 150 source frames.

\subsection{Performance of SLGTformer}
Table~\ref{ablation} presents the ablation studies of the proposed SLGTformer. Compared to vanilla SA along both spatial and temporal dimensions, SLGTformer provides tangible performance uplifts. LGRPE shows a minor boost in performance, while the post-attention graph convolution factor appears to encode node position the best, which is expected since it also contains node self-connections, which boosts temporal performance. TTSA shows both accuracy and efficiency improvements, confirming the better modeling of both short-range and long-range temporal dependencies through LTA and GSTA respectively.

\subsection{Performance on World-Level American Sign Language}
We evaluate SLGTformer performance against non-ensemble RGB, non-ensemble keypoint, and ensemble RGB and keypoint modality based models. Since SLGTformer is proposed as an attention-based backbone to SLR, we can assume that performance improvements propagate through ensemble methods such as \citet{sam_slr_v2}. Compared with other state-of-the-art methods, the proposed SLGTformer achieves the best top-1 and top-5 recognition rates in the non-ensemble joint keypoint modality as shown in Table~\ref{performance}.
\subsection{Limitations}

The offline full-body pose estimator may fail due to frame clipping or occlusion, especially for fingers, which is a common issue for pose estimation
methods. However, fingers play a critical role in expressing
signs. Some of those failures can be corrected by RGB based features, which is why multi-modal ensemble
significantly boosts the recognition rate. Moreover, the same
sign may be performed very differently by different signers
(e.g., signers may use their left hands, right hands, or both
hands to perform the same gloss). Thus, mirroring is important in model training, but more data featuring more signers is ideal. Last, there are visually similar signs (such as heavyweight and lightweight).
Distinguishing between those signs requires very delicate modeling
of spatiotemporal dynamics.

\section{Conclusion}
In conclusion, we propose a novel, attention-based framework, SLGTformer, to learn keypoint sequence feature representations of sign glosses, furthering the robustness and effectiveness of isolated SLR. To our knowledge, there is no existing transformer centric approach to SLR on the keypoint modality that is competitive with the state-of-the-art. SLGTformer is currently the most effective non-ensemble method of modeling motion dynamics from keypoint sequences. 

Specifically, we extract keypoint sequences from RGB video with a pretrained wholebody pose estimator and construct reduced 2D graphs. We propose novel LGRPE as a new attention-based component to encode inter-node neighborhood context from the given graph, while demonstrating the continued effectiveness of the graph convolution factor applied post-attention. Besides the modeling of spatial dynamics via graphs, we propose TTSA as a local-global attention mechanism comprising of LTA and GSTA. This novel one-dimensional temporal attention mechanism uses both windowed self-attention and sub-sampled global self-attention to model short-range and long-range dependencies better, while lowering the complexity of the model in comparison to traditionally quadratic vanilla self-attention time complexity.

SLGTformer is a pioneering project in attention-based SLR, and the potential for transformers to revolutionize the SLR field has yet to be fully realized. We hope our work will facilitate and inspire future SLR research, especially in the continuously broadening field of transformers and dot-product attention as an alternative approach to convolutions in computer vision.

\newpage

\bibliography{iclr2023_conference}
\bibliographystyle{iclr2023_conference}

\end{document}